\documentclass[review,authoryear]{elsarticle}
\usepackage{framed,multirow}
\usepackage{amsmath,amssymb,amsfonts}
\usepackage{latexsym}
\usepackage{algorithmic}
\usepackage{textcomp}
\usepackage{booktabs}          
\usepackage{nicefrac}       
\usepackage{graphicx,wrapfig}
\graphicspath{{Figures/}}
\usepackage{subfig}
\usepackage{tikz}
\usepackage{comment}
\usepackage{multirow}

\DeclareMathOperator*{\argmin}{arg\,min}

\usepackage{xcolor}
\usepackage{url}
\usepackage{hyperref}

\begin{document}
\begin{frontmatter}

\title{A self-attention model for robust rigid slice-to-volume registration of functional MRI }%

\author[1]{Samah Khawaled\corref{cor1}}
\cortext[cor1]{Corresponding author}
\ead{ssamahkh@campus.technion.ac.il}
\author[3,4]{Simon K. Warfield}
\ead{simon.warfield@childrens.harvard.edu}
\author[2]{Moti Freiman \corref{cor2}}
\ead{moti.freiman@technion.ac.il}

\address[1]{The Interdisciplinary program in Applied Mathematics, Faculty of Mathematics, Technion – Israel Institute of Technology}
\address[2]{the Faculty of Biomedical Engineering, Technion – Israel Institute of Technology }
\address[3]{Computational Radiology Laboratory, Boston Children’s Hospital, Boston, MA, USA}
\address[4]{Harvard Medical School, Boston, MA, USA}

\begin{abstract}
Functional Magnetic Resonance Imaging (fMRI) is vital in neuroscience, enabling investigations into brain disorders, treatment monitoring, and brain function mapping. However, head motion during fMRI scans, occurring between shots of slice acquisition, can result in distortion, biased analyses, and increased costs due to the need for scan repetitions.
Therefore, retrospective slice-level motion correction through slice-to-volume registration (SVR) is crucial.
Previous studies have utilized deep learning (DL) based models to address the SVR task; however, they overlooked the uncertainty stemming from the input stack of slices and did not assign weighting or scoring to each slice. In this work, we introduce an end-to-end SVR model for aligning 2D fMRI slices with a 3D reference volume, incorporating a self-attention mechanism to enhance robustness against input data variations and uncertainties. It utilizes independent slice and volume encoders and a self-attention module to assign pixel-wise scores for each slice.
We conducted evaluation experiments on 200 images involving synthetic rigid motion generated from 27 subjects belonging to the test set, from the publicly available Healthy Brain Network (HBN) dataset. Our experimental results demonstrate that our model achieves competitive performance in terms of alignment accuracy compared to state-of-the-art deep learning-based methods (Euclidean distance of $0.93$ [mm] vs. $1.86$ [mm]). Furthermore, our approach exhibits significantly faster registration speed compared to conventional iterative methods ($0.096$ sec. vs. $1.17$ sec.). Our end-to-end SVR model facilitates real-time head motion tracking during fMRI acquisition, ensuring reliability and robustness against uncertainties in inputs. source code, which includes the training and evaluations, will be available soon. 
%Although we trained the model in a fully supervised fashion, it has potential for extension into an unsupervised framework, making it conducive for real-time applications. This novel slice-to-volume registration model presents promising opportunities for improving the quality and efficiency of fMRI data processing, particularly in scenarios where motion correction is crucial for reliable results.
\end{abstract}


\end{frontmatter}

\section{Introduction}
\label{sec:introduction}
Functional MRI (fMRI) is the modality of choice for investigating the functional organization of both healthy and diseased brains \cite{jolles2011comprehensive,fair2008maturing,slichter2013principles}. It plays a pivotal role in neuroscience research and clinical practice, with widespread applications including investigations into brain disorders, treatment monitoring, and brain function mapping, to name a few \cite{slichter2013principles,logothetis2001neurophysiological}. These diverse applications highlight its significance in advancing neuroscience and enhancing patient care.
 
Typically, fMRI data are collected in the form of time series volumes, each comprising stacks of 2D gradient-echo, echo-planar imaging (EPI) slices \cite{stehling1991echo}. These slices are responsible for generating a 3D MRI image that represents the blood oxygenation level-dependent (BOLD) contrast.  fMRI scans, typically lasting from 30 minutes to over an hour, are sensitive to head motion, even with cooperative adult participants \cite{van2012influence}. Head motion can occur at any time during the different shots of the slices acquisition of fMRI, making it susceptible to slice-level, or intra-volume, motion artifacts. This can lead to misalignment of the stack of slices within the overall volume, leading to distortion, biased analyses, and increased costs from scan repetitions \cite{kim1999motion,ciric2017benchmarking}.
In resting state fMRI alone, motion can increase scan times and associated costs by over $57\%$ \cite{dosenbach2017real}. Additionally, a substantial portion of the fMRI data was rendered unusable due to motion, specifically when frame displacement exceeded a predefined threshold (frame displacement$>0.2 mm$) \cite{dosenbach2017real}.
%Head motion represents a notable challenge in fMRI investigations, as highlighted in prior studies \cite{hajnal1994artifacts,friston1996movement}. 
%Additionally, a substantial portion of the fMRI data was rendered unusable due to motion, specifically when frame displacement exceeded a predefined threshold (frame displacement$>0.2 mm$) \cite{dosenbach2017real}.% The substantial rate of data loss, exceeding $50\%$ in pediatric fMRI, underscores the benefits of motion monitoring during data acquisition \cite{dosenbach2017real}. 

Such slice-level head motion highlights the need for retrospecitive slice-to-volume (SVR) registration. Previous studies have demonstrated the superiority of retrospective slice-wise motion correction models for fMRI data over volume-level correction \cite{neves2023real,beall2014simpace,marami2017temporal}.
Real-time motion monitoring empowers scanner operators to promptly intervene, offer feedback to subjects exhibiting motion, and extend data acquisition until a satisfactory amount of motion-free data is obtained \cite{sui2020slimm}. Additionally, it facilitates the development of adaptive scanning strategies based on motion analysis, thereby enhancing the quality of fMRI data \cite{sui2020slimm}.

Traditional approaches have tackled the SVR task by optimizing rigid transformation parameters based on a similarity metric  \cite{gholipour2010robust,kuklisova2012reconstruction,sui2020slimm,kainz2015fast}.  This metric assesses the alignment between the slices and the reference volume, guiding the optimization process. Commonly used metrics include Mean Squared Error (MSE), Cross-correlation, and Mutual Information \cite{maes1997multimodality}. However, iterative optimization methods are time-consuming for real-time intervention. Moreover, some classical methods require coarse alignment of slices as an initialization step, and the quality of the reconstructed volume heavily relies on this initial alignment.

In recent developments, deep learning (DL) techniques have been successfully used to accelerate SVR \cite{salehi2018real,hou20183,xu2022svort,guo2021end}. These DL-based methods predict rigid transformations for MRI slices using Convolutional Neural Networks (CNNs) \cite{kendall2015posenet,guo2021end}.
In \cite{guo2021end}, an end-to-end slice-to-volume registration
network, named as FVR-Net, is designed. 
It estimates the transformation parameters that best align the two images using a dual-branch balanced feature extraction network, which encapsulates both the frame and volume information. The slice encoder of FVR-Net consists of a limited number of layers designed to extract features from an ultrasound frame. This shallowness in the architecture can limit its ability to capture deep features, especially when dealing with a stack of slices instead of a single frame. 
Recently, Transformers \cite{vaswani2017attention} models were trained on synthetically transformed data, to map multiple stacks of fetal MRI slices into a canonical 3D space \cite{xu2022svort}. Such DL-based SVR methods enhanced the capability to handle a broader range of alignments, i.e. transformations that correspond to larger motion \cite{salehi2018real,hou20183,xu2022svort,guo2021end} 

Neither of the aforementioned approaches takes into account the \textit{Aleatoric} uncertainty \cite{kendall2017uncertainties}, which arises from noise in the data or variations within the slices in the stack. Furthermore, they do not allocate weighting or scoring to each slice.
While these input slices share spatial information, it is important to note that not all of them contribute equally to the estimation of the rigid transformation. Some slices contain more valuable information and play a more substantial role in the alignment process. Assigning weights to each slice within the stack offers a straightforward method to account for \textit{Aleatoric} uncertainty.

In light of the aforementioned, we account for the Aleatoric uncertainty in SVR by introducing a self-attention module for DL-based SVR. Our model aligns a stack of multiple slices obtained in a single shot during the fMRI scan with a 3D reference volume, predicting the rigid transformations needed for alignment. It combines dual-branch encoders with self-attention layers to generate a score map for each input slice. Our experiments, conducted with synthetic rigid motion generated on images from the Healthy Brain Network (HBN) dataset \cite{alexander2017open}, demonstrate its competitive performance (the Euclidean distance between the grids sampled with the ground truth and predicted rigid transformations is $0.93$ [mm] compared to $1.86$ [mm]) and quicker run-time ($0.096$ sec. vs. $1.17$ sec.) compared to other existing methods. Source code, which includes the training and evaluations, will be available soon. 

\section{Problem Definition}
\label{sec:problemdef}
 During the fMRI scan, patients may move their heads within the slice acquisition process, resulting in intra-volume or slice-level motion. Head motion is typically modeled as a rigid transformation, characterized by rotations and
translations of acquisition planes, with $6$ degrees of freedom in the 3D canonical space. 
SVR is the process of mapping a single slice or a stack of 2D slices (e.g., fMRI slices acquired in the same acquisition shot during the scan) and a reference 3D volume onto one coordinate system.

SVR can be formulated as an optimization problem. Let us denote the stack of the slices and the reference 3D volume by $\left\{S_k\right\}_{k=1}^{K}$ and $V$, respectively. $T_\alpha$ is the rigid transformation, parameterized by the rotation and translation parameters $\alpha$, which accounts for mapping the grid of the slices to the grid of $V$. Then, we estimate $\alpha$ by minimizing the following energy functional:
\begin{equation}
\hat{\alpha}=\argmin_{\alpha} {L_{Sim}(\left\{S_k\right\}_{k=1}^{K},M\circ T_{\alpha} \circ V)} \label{eq:losseq}
\end{equation}
where $M\circ T \circ V$ denotes the result of warping the moving image with rigid transformation $T_\alpha$ and sampling the corresponding slices according to the input slices indices specified by the operator $M$. $L_{Sim}$ is a dissimilarity, which quantifies the resemblance between the resulting resampled images and the input slices. In CNN-based models, the SVR task is translated to a prediction of the rigid parameters by training the network with the aforementioned dissimilarity loss, $L_{Sim}$. In the training phase, it finds the optimal $\hat{\alpha}$ that maximizes the similarity between the output resampled slices and the input stack.

\section{Methods}
\label{sec:methods}
In the SVR task, the dimensional difference between the 2D and 3D image inputs can significantly impact registration accuracy. A straightforward approach of concatenating or fusing these inputs together can overwhelm the network with volumetric information, often disregarding the valuable details contained within the 2D images.

Instead, inspired by \cite{guo2021end}, our approach employs dual-branch encoders to independently extract features from the slices and the volume. This strategy allows for a balanced utilization of data information, ensuring that the content from the 2D slices is not overshadowed by the volumetric spatial information. Subsequently, these features are concatenated in a late-stage fusion step and fed into a regression model responsible for predicting the rigid parameters. This approach strikes a crucial balance between capturing relevant slice content and incorporating volumetric spatial context, contributing to improved registration accuracy.
To enhance the model's reliability and decrease its sensitivity to Aleatoric uncertainty, we develop a Self-Attention based SVR (SA-SVR) by implementing a mechanism where a pixel-wise score is assigned to each slice. These scores are predicted using a self-attention transformer unit.

\begin{figure}[t!]
\centering
\includegraphics[width=12cm]{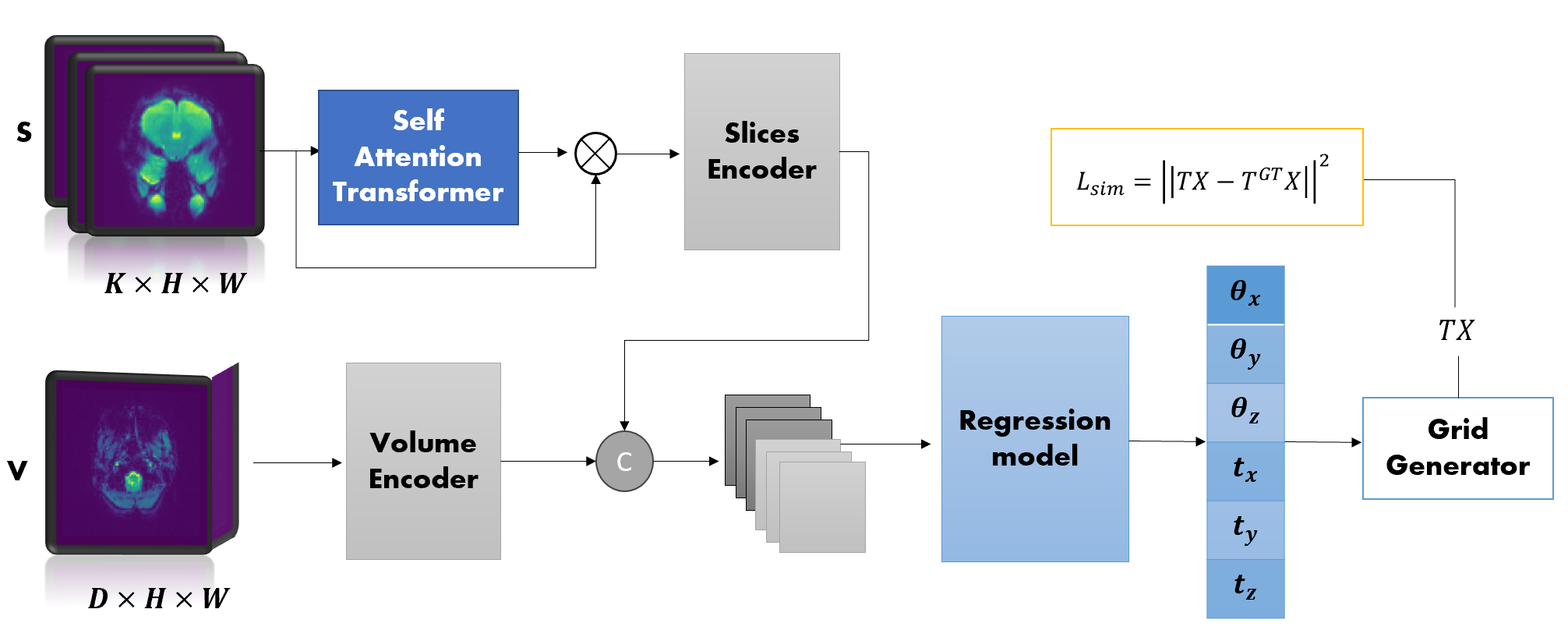}
\caption{Block diagram depicting our SA-SVR system. A self-attention transformer predicts a score for each slice in the stack, followed by the multiplication of each slice by its respective score. Features extracted from both the dual-branch slices and volume encoders are concatenated and employed for the prediction of rigid parameters. }\label{fig:system}
\end{figure}

 Fig.~\ref{fig:system} illustrates the operation of our SA-SVR on a pair of stack of $k$ 2D slices, $S$, each of size $H\times W$, and a reference volume, $V$ of size $D\times H\times W$. Firstly, the self-attention transformer predicts a pixel-wise score for each slice within the stack. Subsequently, these slices are multiplied by their corresponding scores, resulting in weighted slice inputs that are then forwarded to the slices' encoder unit. The dual-branch encoders independently extract features from both the weighted slices and the volumetric data. These features are subsequently fused through concatenation and fed into the regression unit. This regression unit is responsible for estimating the 6 rigid parameters, denoted as $\alpha = (\alpha_x, \alpha_y, \alpha_z, t_x, t_y, t_z)$, which represent the translations and rotations along the three axes.
Utilizing the network's output, i.e., $\alpha$, a 3D grid is generated to reconstruct the slices and calculate the loss during the training phase. In our framework, the loss function primarily relies on the grid rather than being driven by the intensity levels of the images that we aim to align. This comprehensive process forms the core of our registration model.  
\subsection{The Network Architecture}
\begin{figure}[t!]
\centering
\includegraphics[height=4cm]{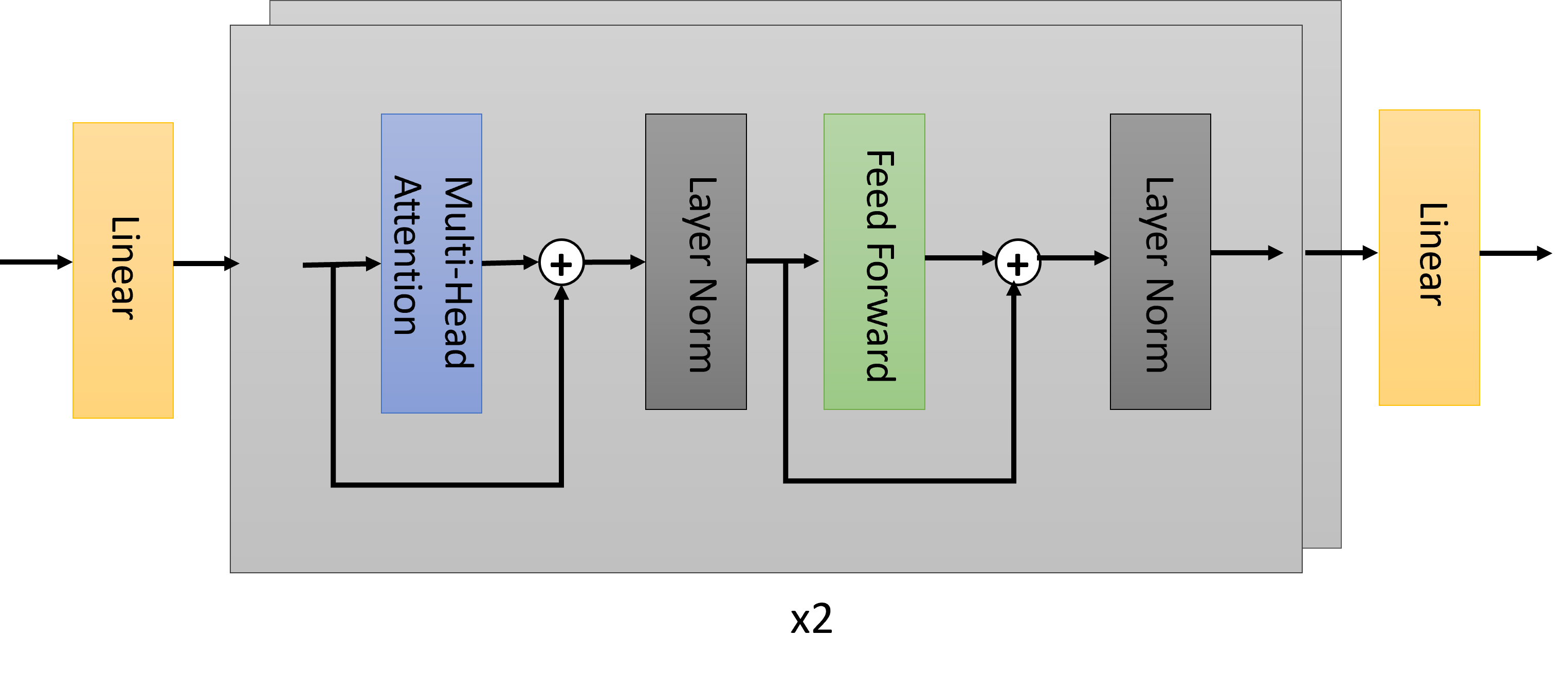}
\caption{A detailed Architecture of the self-attention transformer. The first linear layer embeds the sequence input into a hidden dimension space of $D=256$}\label{fig:transformerarch}
\end{figure}

\paragraph{The Self-Attention Model}
This model is designed for sequence-to-sequence transformations, employing a self-attention mechanism to highlight essential features within the input slices, after they are converted into sequences. This capability empowers it to effectively model long-range dependencies and capture the broader global context.
The self-attention transformer building blocks are described in Fig.~\ref{fig:transformerarch}. It is composed from a linear embedding layer and two consecutive transformer layers, each employing multi-head self-attention and feed-forward sub-layers, followed by output linear layer that maps the transformed output back to the input space. The initial linear layer embeds the sequence input into a hidden dimension space of $256$.  
\paragraph{The Slices and Volume Encoders} 
The slices encoder is composed of a 2D CNN layer that extracts global and low-level features from the input slices while expanding the channel dimension from $K$ to $D$, ensuring match the input volume's size. Following this, a 3D ResNet is applied. Both the slices and volume encoders adopt the ResNet-10 architecture \cite{hara3dcnns}, consisting of four stages, each with a single residual block.
\paragraph{The Regression Model}
The regression model comprises three layers, each consisting of a single ResNeXt block \cite{xie2017aggregated}. Subsequently, average pooling layer is applied. Finally, it is followed by the output linear layer which dedicated to predicting the $6$ rigid parameters, i.e. the rotation angles and translations. 

\subsection{Representations of Transformations and Slices Generation}\label{subsec:transgen}
\begin{figure}[t!]
\begin{minipage}[b]{0.45\linewidth}
    \centering
    \subfloat[\label{fig1:a}\footnotesize Coronal  ]{
    \includegraphics[width=6cm]{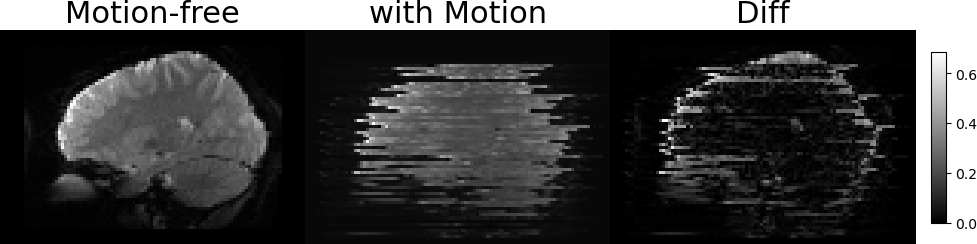}}
\end{minipage}
\begin{minipage}[b]{0.45\linewidth}
    \centering
    \subfloat[\label{fig1:b}\footnotesize Coronal ]{
    \includegraphics[width=6cm]{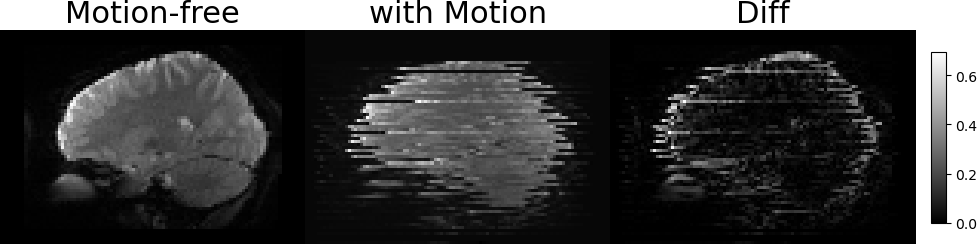}}
\end{minipage} \\
\begin{minipage}[b]{0.45\linewidth}
    \centering
    \subfloat[\label{fig1:c}\footnotesize Sagittal ]{
    \includegraphics[width=6cm]{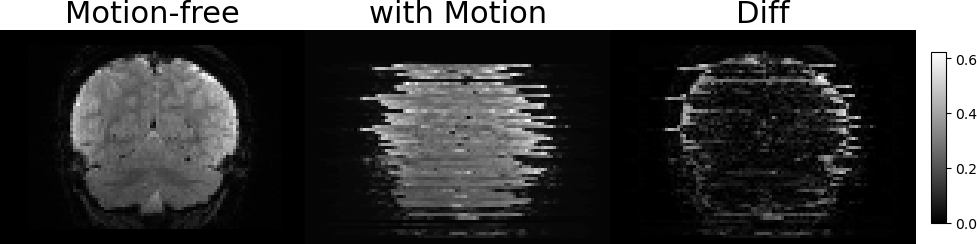}}
\end{minipage} 
\begin{minipage}[b]{0.45\linewidth}
    \centering
    \subfloat[\label{fig1:d}\footnotesize Sagittal ]{
    \includegraphics[width=6cm]{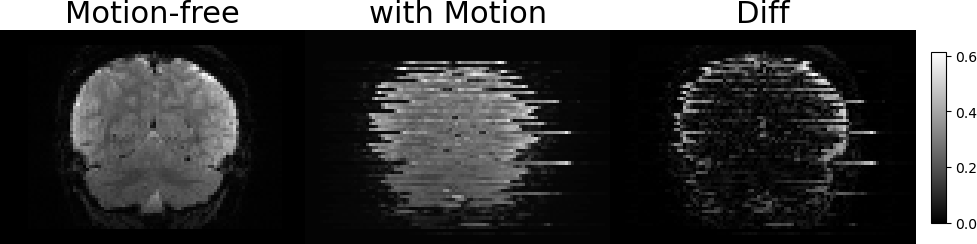}}
\end{minipage} \\
\begin{minipage}[b]{0.45\linewidth}
    \centering
    \subfloat[\label{fig1:e}\footnotesize Axial ]{
    \includegraphics[width=6cm]{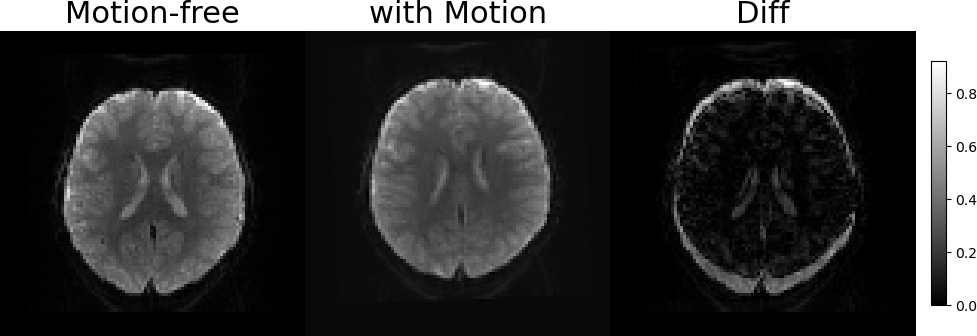}}
\end{minipage} 
\begin{minipage}[b]{0.45\linewidth}
    \centering
    \subfloat[\label{fig1:f}\footnotesize Axial ]{
    \includegraphics[width=6cm]{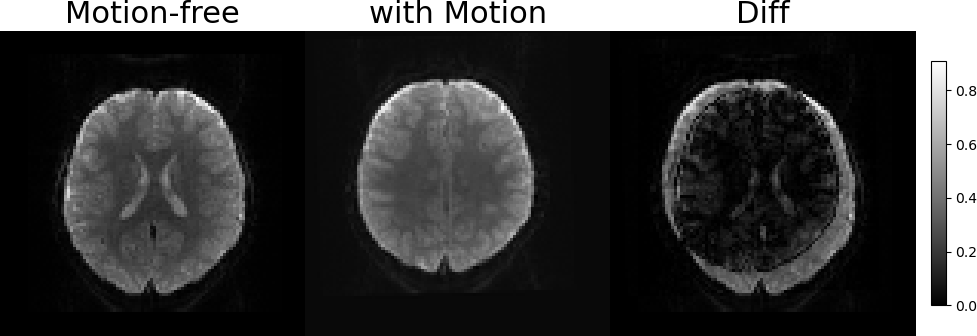}}
\end{minipage} 
\caption{ Illustration of the SVR problem.  Two examples of free-motion volumes and volumes with synthetic motion. From left to right: Coronal \protect\subref{fig1:a},\protect\subref{fig1:b} and Sagittal \protect\subref{fig1:c},\protect\subref{fig1:d} views of the free-motion volumes, the generated volumes after applying the rigid transformations and sampling the slices, and the pixel-wise MSE between them, respectively. The artifacts due to the slice-level motion are visible in both Sagittal and Coronal views, however, they are not dipicted in the axial (imaging) axis of the image \protect\subref{fig1:e},\protect\subref{fig1:f}.  }\label{fig:motionexample}
\end{figure}

Supervised learning with our model demands knowledge of the ground truth transformation for each stack of slices. However, manually annotating the precise 3D location of 2D MRI slices poses a significant challenge. As an alternative, we employ artificially sampled 2D slices from high-quality fMRI volumes with minimal head motion. To encompass a broad spectrum of rigid transformations, we uniformly sample rotation and translation parameters within the ranges specified in Table \ref{tab:ranges}. 
Beginning with the given rotation angles often represented as Euler angles, we construct the rotation matrix as a product of sequential rotations around distinct axes, such as roll, pitch, and yaw. Affine transformation matrices are then derived by merging the rotation matrix with translation vectors, capturing both the change in orientation and the spatial displacement in the 3D space. 
 These matrices collectively constitute an affine transformation, which is subsequently applied to each reference 3D volume input.  
Finally, a stack of 2D slices is sampled from these volumes based on a predefined protocol. This protocol is related to the acquisition protocol of the fMRI database, which specifies the slices indices, and how and when they were collected. In our settings, the number of 2D slices in each stack is $K=6$, according to the fMRI acquisition protocol $6$ slices were simultaneously acquired. Fig.~\ref{fig:motionexample} showcases examples of free-motion volumes and volumes with synthetically generated motion following the described process. Each stack of slices is sampled from volumes with motion, each parameterized with distinct parameters. This process leads to visible discontinuities, as depicted in the sagittal and coronal views of the volumes featuring motion. This emphasizes the inter-volume motion, which varies from one stack to another.   

\subsection{Loss Function}
Our SA-SVR model is trained with fully supervised fashion, where it directly uses
ground-truth rigid parameters to constraint the estimation. The overall loss term is:
\begin{equation}
    L = L_{sim} + \lambda_1 L_{ang} + \lambda_2 L_{tr}
\end{equation}
where $L_sim$ is the dissimilarity loss defined by the following Euclidean distance:
\begin{equation}
    L_{sim} = \Vert T^{GT}X-T^{\hat{\alpha}}X \Vert_2  
\end{equation}
where $T^{\hat{\alpha}}$, $T^{GT}$ are the affine matrices generated by the predicted and ground truth rigid parameters. $X$ is the 3D grid. The Euclidean distance between the transformation matrices indirectly measures the dissimilarity between the transformed grids. $ L_{ang}$ and $L_{tr}$ are the Mean-squared-error (MSE) between the predicted and ground truth rotation angles and translation, respectively. $\lambda_1$ and $\lambda_2$ are tuning hyper-parameters that balance between the two aforementioned expressions.

\begin{table}[]
\centering
\resizebox{0.5\textwidth}{!}{%
\begin{tabular}{llll}
                      & x (Pitch)  & y (roll)  & z (Yaw)  \\ \hline
translations {[}mm{]} & $\left [-12,12\right]$ & $\left [-12,12\right]$ & $\left [-8.4,8.4\right]$ \\
rotations {[}deg{]}   & $\left [-5,5\right]$  & $\left [-5,5\right]$  & $\left [-5,5\right]$  
\end{tabular}%
}\caption{Ranges of Rigid Parameters. The ground truth rigid parameters are sampled from a uniform distribution in the ranges $(-a,a)$, where $a$ (the maximal value) is specified for each parameter. } \label{tab:ranges}
\end{table}

\section{Experimental Methodology}
\paragraph{Database and Preprocessing}
To train and evaluate our model on large real pediatric data, we used the Healthy Brain Network (HBN) dataset \cite{alexander2017open}. This database contains resting state fMRI scans of $138$ subjects. The age range of these subjects is from $5.8$ to $21.4$ years. The fMRI time series have $375$ measurements, slice thickness of $2.4 mm$ (isotropic spacing), $60$ slices, and matrix size of $84 \times 84$. 
for each scan, the fMRI volume at time $t=0$ was selected as reference volumes. In total, we have $138$ volumes. All images were then zero-padded to a size of $70\times 100\times 100$, and normalized to the $\left[0,1\right]$ gray-scale domain. 
We split these volumes into training, test, and validation sets with the following ratios: $64\%$, $20\%$, and $16\%$, respectively.
Then, from these, we generated  $2000$, $500$, and $200$ pairs of 3D volumes and stacks of 2D slices for training, validation, and testing, respectively. The rigid transformations for each set were synthesized as described in \ref{subsec:transgen} and the slices indices were sampled as the sampling protocol of the HBN database. In the HBN fMRI acquisitions, $6$ slices are simultaneously acquired and the slice order of these is $(i,i+j_1,...,i+j_5)$, where $j_k = k \times \frac{n}{6}$ and $i \in \left\{1,2,...,n \right\}$, and $n$ is the number of the overall slices in the 3D volume.

\paragraph{Implementation Details}
We implemented our method for SVR of fMRI images using Pytorch backend \cite{paszke2017automatic}.%\footnote{source code, which includes the training and evaluation will be available soon. } 
The number of heads used in the self-attention layer was empirically set to $H=8$. However, during training, we experimented with different values of $H$, specifically $H=4$ and $H=16$, and found that using $H=8$ resulted in reduced loss on the validation set. The affine grid resampling and the interpolation were implemented in Pytorch as well \cite{paszke2017automatic}
The SA-SVR network was trained for $300$ iterations and with batch size equal to $8$. We used an \texttt{Adam} optimizer \cite{kingma2014adam} with  a learning rate set to $5\times10^{-4}$. 
We conducted an exhaustive search to determine the best value for the tuning parameters, $\lambda_1$ and $\lambda_2$, which provides best performance in terms of registration accuracy.
\paragraph{fMRI Motion Simulation Study}
In this study, we evaluate the prediction accuracy of the motion correction achieved by our SVR approach during fMRI acquisition. Due to the absence of real motion parameters, we simulate rigid motion. In this experimental setup, we initially select reference volumes characterized by minimal head motion. For each subject, we apply VVR method, provided by ANTs \cite{avants2011reproducible}, to align each time point's volume with the reference volume scanned at $t=0$. Subsequently, we identify the $10$ volumes with the least $\Vert A-I \Vert$, where $A$ represents the affine matrix aligning each volume with the reference. From these selected volumes, we generate volumes with simulated rigid motion and employ our SA-SVR method to register them to the reference volume.

\paragraph{Evaluation Methodology}
We evaluate the accuracy of the SVR by measuring the following metrics: (1) The distance error, $D_{reg}$, which denotes the average distance in
millimeters between the grids sampled with the ground truth and predicted rigid transformations. (2) The rotations and translations errors $E_{rot}$ and $E_{tr}$, which measures the MSE between the estimated and ground truth rotation and translation parameters.  
we trained our SA-SVR with multiple settings of the loss by changing the parameters $\lambda_1$ and $\lambda_2$.
We compare our performance with FVR-Net, a state-of-the-art DL-based model for frame-to-volume registration. We train FVR-Net with the same settings and loss as described in \cite{guo2021end}, while modifying the slices input to be a stack of $6$ slices instead of a single frame.  Additionally, to assess the added value of the self-attention module, we compare our SA-SVR with a baseline model, which has identical architecture and loss without the self-attention module. We train the baseline SVR with the same settings and optimizer. 

\section{Results and Discussion}
%%%% evaluation metrics 
 Table~\ref{tab:results_tune} presents the registration results of the SA-SVR model for 3 different values of $\lambda_1$ and $\lambda_2$ by means of registration accuracy measured on the overall test set. 
We selected the model with $\lambda_1=10$ and $\lambda_2=100$ which optimized both the translations and rotations estimation error ad balanced the three terms in our loss.   

%%%% Comparision with other methods 
Table~\ref{tab:results_eval} presents the averaged values of the evaluation metrics, the grid distance, and the rigid parameters errors, over the whole test set.  Our SA-SVR achieved improvements over FVR-Net in terms of registration accuracy (a paired test with a p-value of $p<0.03$ for both rotation and translations MSEs). It is worth mentioning that our comparison with well-known conventional optimization methods for SVR such as ANts and SVRTK didn't apply well to our task settings due to the large translations in the synthetic motion.   

Table~\ref{tab:runtime_memory} provides a summary of the performance comparison between our method and FVR-Net in terms of both runtime and complexity. Our method demonstrates a faster average registration time for pairs of slices and volumes compared to FVR-Net. Additionally, our network has significantly fewer trainable parameters in comparison to FVR-Net, making it less resource-intensive for training and evaluation. 

\begin{figure}[t!]
\begin{minipage}[b]{0.32\linewidth}
    \centering
    \includegraphics[width=4cm]{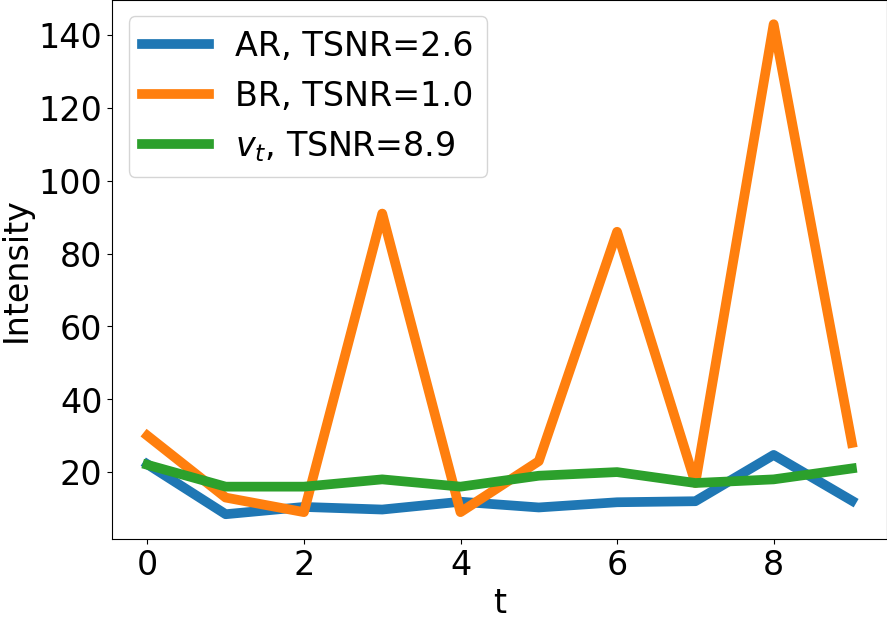}
\end{minipage}
\begin{minipage}[b]{0.32\linewidth}
    \centering
    \includegraphics[width=4cm]{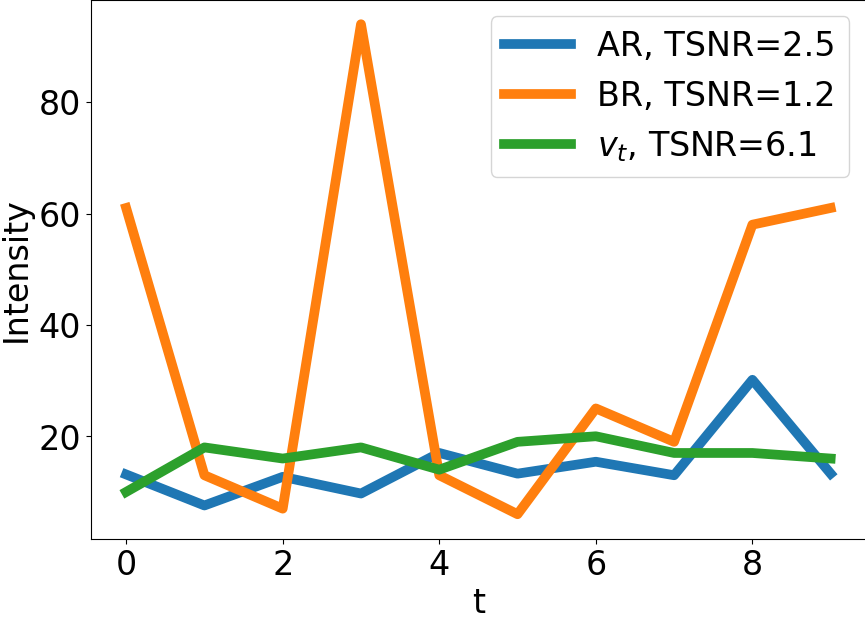}
\end{minipage}
\begin{minipage}[b]{0.32\linewidth}
    \centering
    \includegraphics[width=4cm]{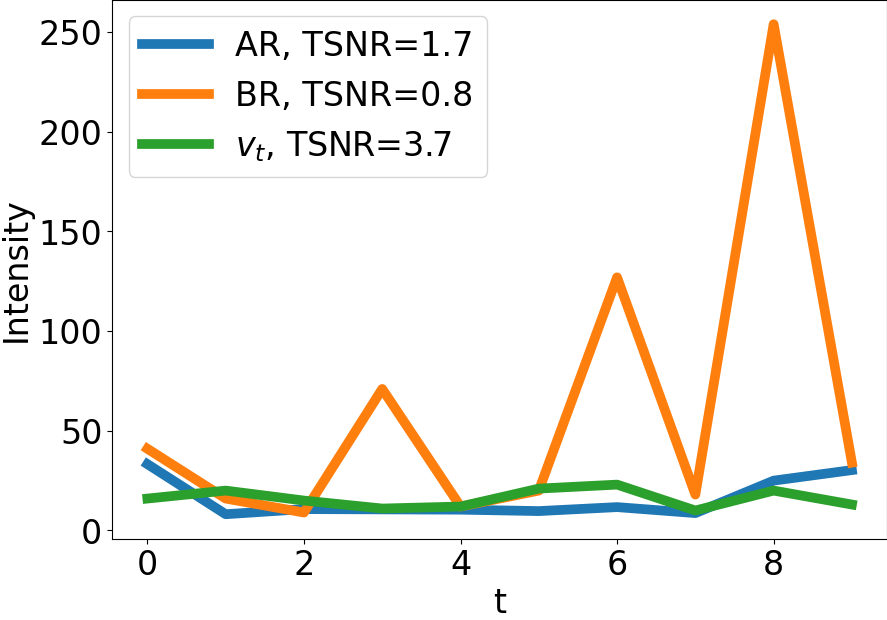}
\end{minipage} \\ 
\begin{minipage}[b]{0.32\linewidth}
    \centering
    \includegraphics[width=4cm]{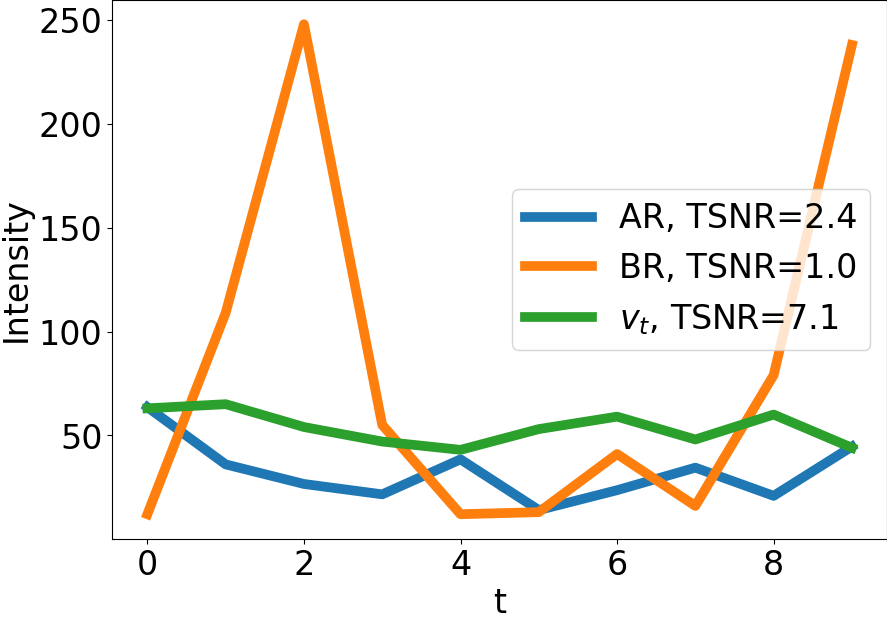}
\end{minipage}
\begin{minipage}[b]{0.32\linewidth}
    \centering
    \includegraphics[width=4cm]{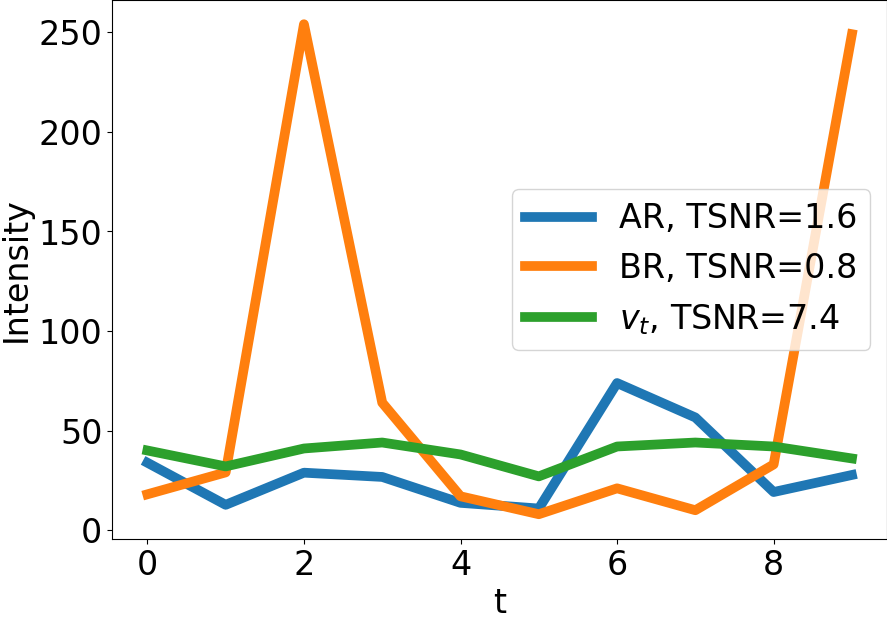}
\end{minipage}
\begin{minipage}[b]{0.32\linewidth}
    \centering
    \includegraphics[width=4cm]{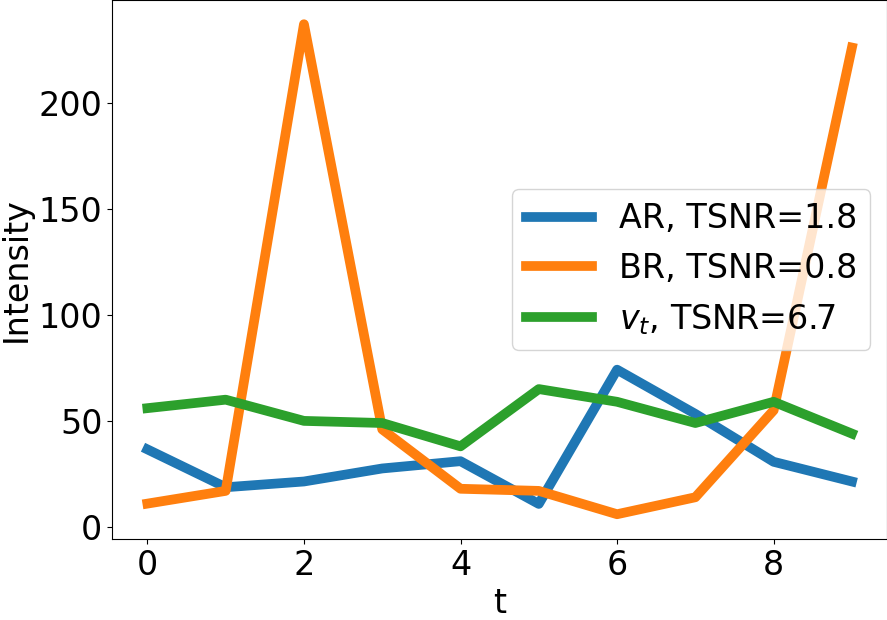}
\end{minipage}

\caption{fMRI motion-correction results. Sampled intensity values from various voxels across the registered volume (marked as AR - after registration), the motion-free volume ($V_t$), and the volume with simulated motion (denoted as BR - before registration), collected over time points representing the course of the fMRI series. 
The voxels used in the analysis for the upper and bottom parts were sampled from the left and right red boxes in Fig.~\ref{fig:pixelrois}, respectively.}\label{fig:pixelfmri}
\end{figure}

\begin{figure}[t!]
\begin{minipage}[b]{0.45\linewidth}
    \centering
    \includegraphics[width=4cm]{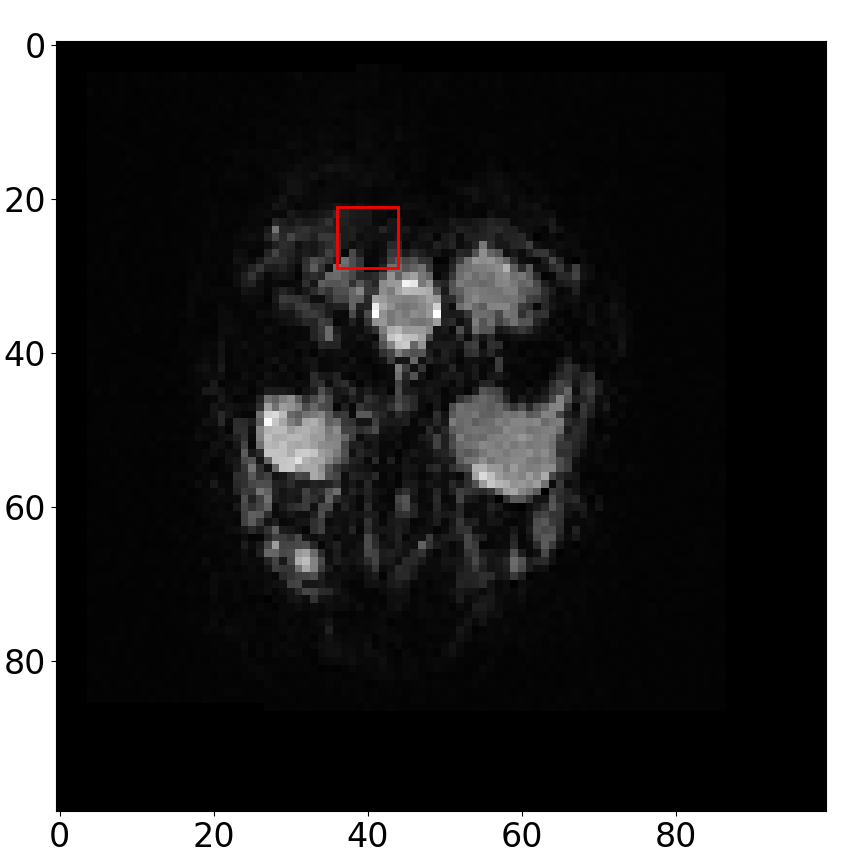}
\end{minipage}
\begin{minipage}[b]{0.45\linewidth}
    \centering
    \includegraphics[width=4cm]{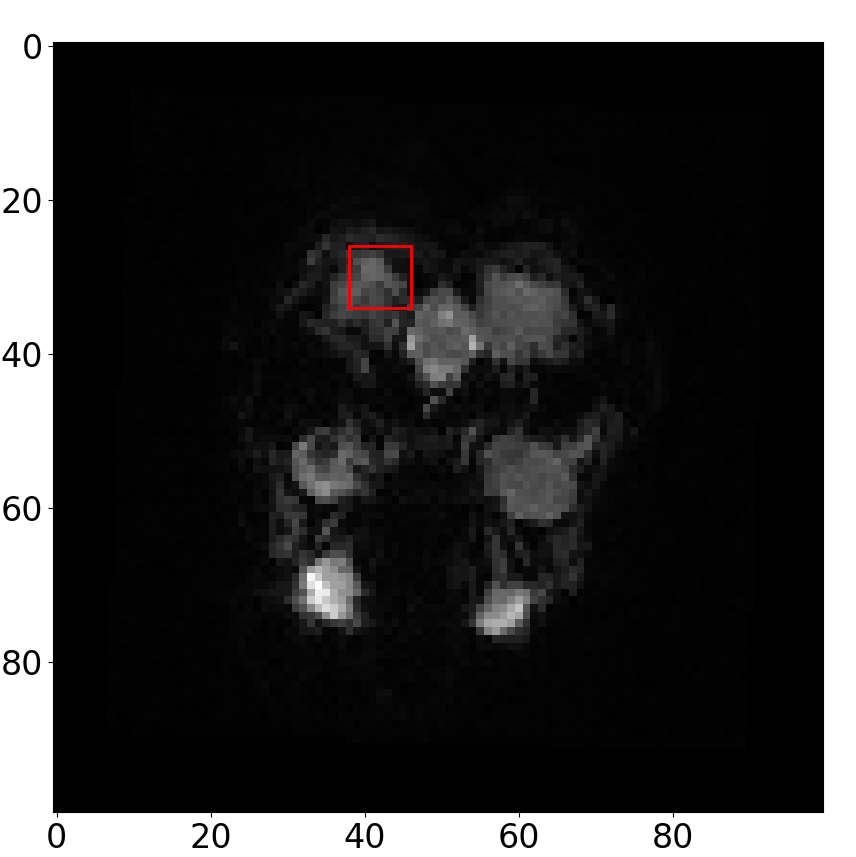}
\end{minipage}
\caption{ROIs of the selected pixels. From the highlighted red regions, we sampled voxels used for the fMRI motion correction experiment. The upper and bottom figures in Fig.~\ref{fig:pixelfmri} show results on voxels sampled from the left and right red boxes, respectively.}\label{fig:pixelrois}
\end{figure}

Fig.~\ref{fig:pixelfmri} depicts the results of the fMRI motion correction study. To illustrate this, we computed intensity values at specific voxels both before and after applying our SVR method, and we present these values over the time points that span the course of the fMRI series, where synthetic motion was introduced. 
The regions of interest (ROIs) from which we sampled voxels are depicted in Fig.~\ref{fig:pixelrois}.
As observed, the intensity values after motion correction exhibit reduced variations compared to those before alignment and are more closely aligned with the intensity values of the reference volume.

\begin{table}[t]
\centering
\resizebox{0.9\textwidth}{!}{%
\begin{tabular}{c|c|c|c|c|c}
$\lambda_1$ & $\lambda_2$ & $D_{init} (mm)$         & $D_{reg}(mm)$   & $E_{rot}(deg)$     & $E_{tr} (mm)$     \\ \hline \hline
$1$           & $100$         & $12.99\pm2.75$ & $0.92\pm0.52$       &  $0.21\pm0.11$   & $0.30\pm0.21$ \\
$10$          & $40$          & $12.99\pm2.75$ & $1.12\pm0.88$   &  $0.097\pm0.053$ & $0.42\pm0.32$ \\
$10$          & $100$      & $12.99\pm2.75$ & $0.93\pm0.59$ &  $0.12\pm0.057$  & $0.35\pm0.22$
\end{tabular}%
}
\caption{Hyper-parameters tuning results.The mean and std values of the three evaluation metrics, calculated over the whole test set, for three values of $\lambda_1$ and $\lambda_2$. $D_{init}$ is the distance error before applying the registration. } \label{tab:results_tune}
\end{table}

\begin{table}[t]
\centering
\resizebox{0.9\textwidth}{!}{%
\begin{tabular}{c|c|c|c|c}
 & $D_{init}(mm)$   & $D_{reg}(mm)$   & $E_{rot} (deg)$     & $E_{tr} (mm)$    \\ \hline \hline
SA-SVR  & $12.99\pm2.75$ & $\mathbf{0.93\pm0.59}$ & $0.12\pm0.057$ & $\mathbf{0.35\pm0.22}$ \\
Baseline & $12.99\pm2.75$ & $1.18\pm0.72$  & $0.27\pm0.13$   & $0.38\pm0.30$ \\
FVR-Net   & $12.99\pm2.75$   & $1.86\pm0.97$ & $\mathbf{0.11\pm0.05}$   & $0.71\pm0.34$
\end{tabular}%
}
\caption{Registration accuracy evaluation results. The mean and std of the evaluation metrics, calculated over labels and test examples for each database (columns), are presented for the three different models (from top to bottom: SA-SVR, the baseline SVR without the self-attention module, and the benchmark FVR-Net, respectively.}\label{tab:results_eval}
\end{table}

\begin{table}[t]
\centering
\resizebox{0.5\textwidth}{!}{%
\begin{tabular}{c|c|c}
                & \# parameters & Run time (sec) \\ \hline
\textbf{SA-SVR} & $\simeq12.2
\times10^6$          & $0.096$          \\
\textbf{Baseline} & $\simeq3.13\times10^6$          & $0.095$          \\
\textbf{FVR-Net} & $
\simeq29.35\times10^6$         & $1.1750$       
\end{tabular}%
}\caption{Runtime and complexity comparison. }\label{tab:runtime_memory}
\end{table}

\section{Conclusion}
In this paper, we introduce a novel end-to-end slice-to-volume registration model designed to align a stack of 2D fMRI slices with a 3D reference volume in a combined 3D space. Our approach is capable of mimicking real-time scenarios, allowing for instant registration of acquired slices during data acquisition. The model incorporates independent slices and volume encoders, as well as a self-attention module, which predicts pixel-wise scores for each slice in the stack to quantify their contribution to the alignment task.

Our experiments on synthetic rigid motion showcase competitive performance in terms of alignment accuracy and rigid parameter estimation when compared to other state-of-the-art deep learning-based methods. Additionally, our model demonstrates faster registration speed compared to traditional iterative registration methods. While our approach is trained in a fully supervised fashion, its potential extension to an unsupervised framework would facilitate its use in real-time applications. 

\section*{Acknowledgements}
Khawaled, S. is a fellow of the Ariane de Rothschild Women Doctoral Program. This project is funded by the Prof. Rahamimoff Travel Grant for Young Scientists, through the U.S.-Israel Binational Science Foundation (BSF).

\newpage
\bibliographystyle{model2-names.bst}
\biboptions{authoryear}
%\biboptions{els-num}
\bibliography{refs.bib}
\end{document}